\documentclass{egpubl}
\usepackage{arXiv}
 
\WsSubmission      

\usepackage[T1]{fontenc}
\usepackage{dfadobe}  

\usepackage{cite}  
\BibtexOrBiblatex
\electronicVersion
\PrintedOrElectronic
\ifpdf \usepackage[pdftex]{graphicx} \pdfcompresslevel=9
\else \usepackage[dvips]{graphicx} \fi

\usepackage{egweblnk} 
\usepackage{graphicx}
\usepackage{amsmath}
\usepackage{amssymb}
\usepackage{booktabs}

\usepackage{placeins}
\usepackage{algorithm}
\usepackage{algorithmic}
\usepackage{array}
\usepackage{overpic}
\usepackage{multirow, makecell}
\usepackage{bm}
\usepackage{wrapfig}

\usepackage{pgfplots}
\usepackage{pgf,tikz}

\title[Localized Gaussians as Self-Attention Weights for Point Clouds Correspondence]%
      {Localized Gaussians as Self-Attention Weights \\ for Point Clouds Correspondence}

\author[A. Riva, A. Raganato, and S. Melzi]
{
\parbox{\textwidth}{
    \centering
    A. Riva$^{1}$\orcid{0009-0007-0177-0604},
    A. Raganato$^{1}$\orcid{0000-0002-7018-7515},
    and S. Melzi$^{1}$\orcid{0000-0003-2790-9591}
}
\\
{\parbox{\textwidth}{\centering
        $^1$University of Milano-Bicocca, Department of Informatics, Systems and Communication, Italy
    }
}
}

\begin{document}

\teaser{
 \includegraphics[width=\linewidth]{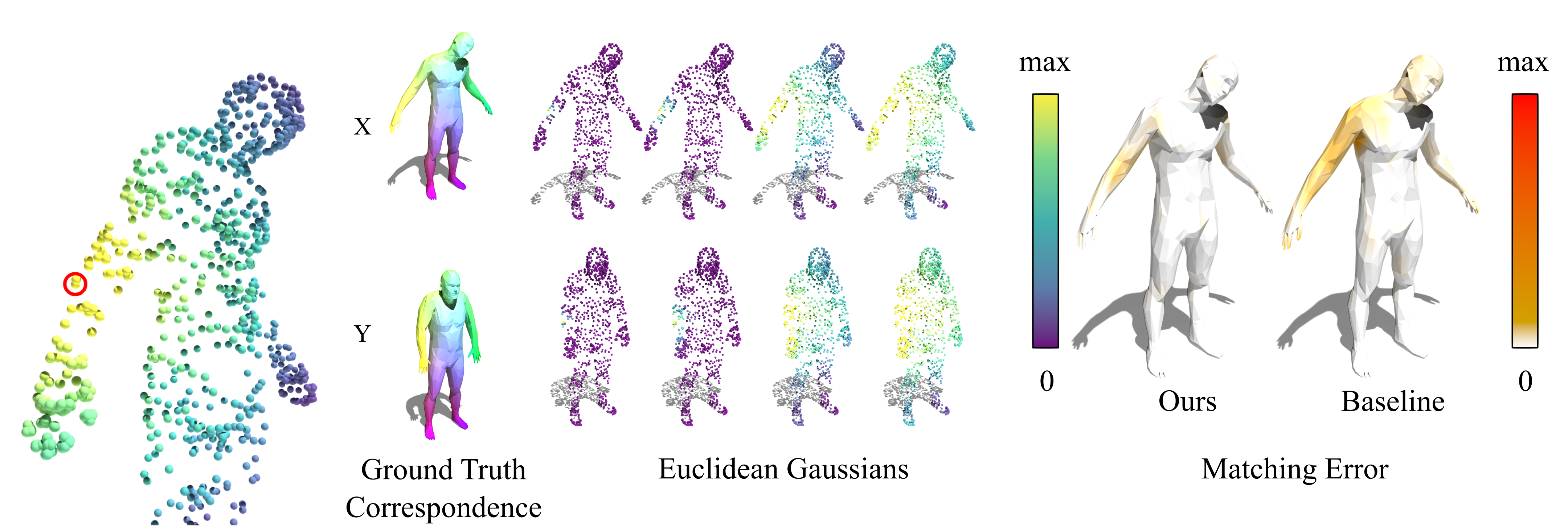}
 \centering
  \caption{On the left, a close-up of the Gaussian function plotted over the point cloud with the center point highlighted in the red circle. In the center, the ground truth for the matching of the points where we depict corresponding points with the same color. On the right, a heatmap of the errors produced by our model compared to the ones of the baseline model.}
\label{fig:teaser}
}

\maketitle
\begin{abstract}
    Current data-driven methodologies for point cloud matching demand extensive training time and computational resources, presenting significant challenges for model deployment and application. In the point cloud matching task, recent advancements with an encoder-only Transformer architecture have revealed the emergence of semantically meaningful patterns in the attention heads, particularly resembling Gaussian functions centered on each point of the input shape.
    In this work, we further investigate this phenomenon by integrating these patterns as fixed attention weights within the attention heads of the Transformer architecture. We evaluate two variants: one utilizing predetermined variance values for the Gaussians, and another where the variance values are treated as learnable parameters. Additionally we analyze the performances on noisy data and explore a possible way to improve robustness to noise.
    Our findings demonstrate that fixing the attention weights not only accelerates the training process but also enhances the stability of the optimization. Furthermore, we conducted an ablation study to identify the specific layers where the infused information is most impactful and to understand the reliance of the network on this information.


\end{abstract}  
\section{Introduction}
\label{sec:intro}
As consumer interest in augmented and mixed reality grows, 3D data acquisition technologies are becoming increasingly accessible. This heightened interest underscores the need for more efficient 3D algorithms to handle real-world imperfections while delivering high-quality results.
Among various applications involving 3D data, human-related tasks are particularly significant due to their central role in human-machine interactions. These tasks present considerable challenges because of the wide variety of human body shapes and sizes. Developing methods that work universally across this diversity is a complex problem, as even seemingly reasonable assumptions often fall short when applied to real-world data \cite{trewin2018ai}.
Human data is often used not only in virtual reality but also in other fields, such as surveillance, autonomous driving, and medical applications, which require 3D reconstructions of patient bodies. These tasks often rely on geometric data and require real-time processing \cite{huang2017review, contreras2024survey}.
Moreover, 3D data inherently presents challenges due to its loosely structured nature. While image and 2D data methods benefit from the well-defined structure of pixel matrices, geometric data lacks such structure and is often sparse.

In recent years, data-driven methods have emerged for various 3D tasks, including object annotation, segmentation, classification, and geometry generation. In this work, we focus on the point cloud matching task, which involves finding correspondences between points discretizing a pair of shapes. This is a fundamental step crucial for 3D registration.
The Transformer architecture has been successfully employed for shape matching and registration tasks, either alone or as part of broader architectures, yielding impressive results \cite{wang2019deep, fu2021robust, trappolini2021shape, lu2022transformers}.
However, while these architectures are relatively lightweight during inference, they are notably resource-intensive during the training phase.
Recently, Raganato et al. \cite{raganato2023attention} observed the formation of Gaussian patterns in the attention weights of Transformers used for point cloud matching. In this work, we further investigate these patterns and explore how they can be exploited to enhance performance and reduce the training time of encoder Transformer models for the matching problem, thereby also reducing the environmental impact of the process.
Our studies demonstrate that injecting Gaussian information into some self-attention heads in Transformer-based models can stabilize optimization, reduce the parameter footprint, and improve correspondence quality.

Our contributions can be summarized as follows:
i) we improve the training process and reduce the parameter count of a Transformer architecture by fixing the weights of some self-attention heads to localized Gaussian functions.
ii) we optimize the parameters of the Gaussians to identify the best configuration for human shapes.
iii) we study the robustness to noise of the methods and explore a way to improve on it.
iv) we analyze the impact of the injected information on different components of the network.
\section{Related work}\label{sec:related}

\subsection{Non-rigid shape matching}
It is fundamental for many applications to map each point of a surface to one point of a second surface that has undergone a non-rigid deformation. For this problem, known as \textit{non-rigid shape matching}, many contributions have been raised with a number of approaches.
Descriptor-based methods long offered a solution by computing a vector of features, invariant to a number of transformations, for each point. Then, the matches are assigned according to a similarity score between points in the feature space \cite{salti2014shot}, \cite{sahillioǧlu2011coarse}, \cite{tevs2011intrinsic}.
A more comprehensive review of the matching methodologies can be found in \cite{deng2022survey} and, in particular, for the methods that extrinsically solve the correspondence problem in \cite{sahilliouglu2020recent}.
While hand-crafted methods for shape correspondence produce good results, they present high developmental costs and are often domain-specific, making it difficult to apply them in other fields.
Thanks to the rise of machine learning, more data-driven methods have been proposed, and the transformer architecture has been proven to be a good solution for this problem and to generalize well to a number of different contexts.
Some methods involving an encoder-decoder architecture, naturally designed to allow interactions between more inputs, have emerged; for example, in \cite{trappolini2021shape}, the Perceiver architecture \cite{jaegle2021perceiver}, originally proposed to target classification tasks across various modalities, is adapted to the shape matching task.
Oppositely, in \cite{raganato2023attention}, the authors proposed a shape matching solution with state-of-the-art competitive performances that exploits an encoder-only transformer architecture.
This work pointed out the importance of introducing positional information about the points and conducted an extensive ablation study to identify the critical points of the architecture.
In \cite{raganato2023attention}, it has also been revealed that the attention patterns produced by the model are close to Gaussians centered around the points. Furthermore, it has been supposed that directly providing these patterns to the network could significantly speed up the learning procedure; in this work, we aim to explore this possibility, the advantages it brings and its drawbacks.

\subsection{Attention in Transformers}
Transformer architectures have been initially introduced to tackle Natural Language Processing (NLP) tasks \cite{vaswani2017attention} and have, over the years, achieved better and better performances also in fields beyond NLP \cite{khan2022transformers,li2023survey}.
The typical input for Transformer-based models is one or more sequences of tokens, in NLP those are often a direct representation of words.
The attention mechanism, which is the core of the Transformer architecture, works by computing weights that encode the relationships between a given token and all other tokens in a sequence. In self-attention, this process is applied within the same sequence, allowing the model to consider the context of each token relative to all others in that sequence. In cross-attention, instead, it is used to relate tokens from different sequences. Multi-head attention enhances this mechanism by computing multiple sets of attention weights, known as heads, in parallel. These heads are then concatenated, enabling the network to capture different relationship aspects independently and learn complex patterns.

Given their success and widespread usage, numerous studies have focused on interpreting Transformer networks, particularly analyzing attention mechanisms and the interpretability of their weights and connections \cite{clark-etal-2019-bert,lyu2024towards,madsen2022post,wang2023interpretability}. These analyses have led to several advancements aimed at improving the network's efficiency and performance. For instance, in the context of machine translation in NLP, multiple studies have shown that certain attention patterns learned by Transformer architectures reflect positional encoding of contextual information \cite{raganato2018analysis,voita-etal-2019-analyzing,rogers2021primer}. These straightforward patterns can be integrated into the architecture without the need for extensive training. \cite{raganato2020fixed} and \cite{you2020hard} significantly simplified the model by replacing some attention heads with fixed self-attention patterns or Gaussian patterns, respectively, which reduces the number of parameters while minimally impacting performance.
A similar solution is reported in \cite{tay2021synthesizer}. The authors propose a model that learns synthetic attention weights without token-token interactions.
By combining synthetic attention heads with dot-product attention heads, the model outperforms the traditional transformer, showing how the attention mechanism can often be improved with the use of less complex functions.
\section{Background}
\label{sec:background}
In this section, we describe the problem addressed in our work and introduce the underlying concepts of the explored methodology.

\subsection{The task}
Broadly speaking, the shape matching task involves identifying a bijective map between the points of two surfaces. When restricted to discrete settings such as point clouds and meshes, this task can be defined as finding a set $\Pi_{X,Y} \subset X \times Y$, where $(x,y) \in \Pi_{X,Y}$ implies that $y$ corresponds to $x$ for each point $x \in X$ and $y \in Y$. In the case of meshes, the problem can incorporate additional constraints due to the connectivity defined by the edges.
In this work, we focus only on unordered point clouds representing human bodies undergoing non-rigid deformations.

\subsection{Attention patterns}
\label{ssec:shape_matching}%
\begin{figure}[h]
    \centering
    \includegraphics[width=\linewidth]{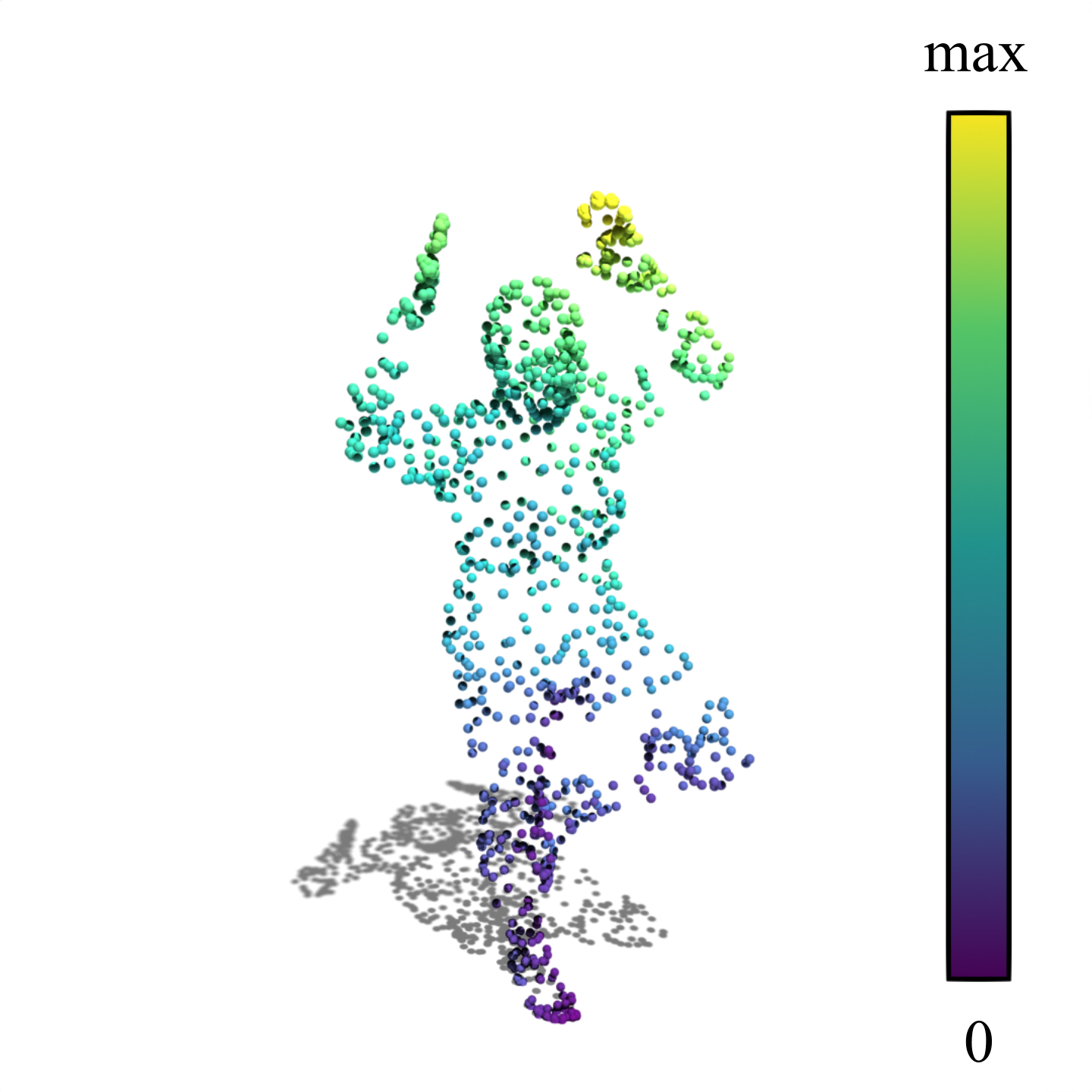}
    \caption{Example of attention pattern in \cite{raganato2023attention} plotted over the shape for one point (top left hand finger). The point colors represent a heatmap with a spike on the left hand (yellow) and small values on the feet (blue).}
    \label{fig:attention_pattern_example}
\end{figure}

In \cite{raganato2023attention}, insights into the functions of attention heads in Transformer models for point cloud matching are provided. The analysis reveals that the model generates attention heads that approximate diagonal blocks in both directions. Given that the input of the model is the concatenation of two point clouds, this suggests that the attention heads specialize in producing self-attention and cross-attention patterns.
Furthermore, these patterns, when plotted as signals over the shapes, resemble Gaussian distributions centered around the points of the shape. This indicates that the model actively retrieves a neighborhood around each point to encode relevant information about it.
Figure \ref{fig:attention_pattern_example} illustrates one such pattern (half of a row in an attention head from \cite{raganato2023attention}) for a point on the left hand. It is evident that the attention weights are larger around the left hand (yellow) and decrease as the points get farther from the target point, with values close to zero in the farthest regions, like the feet (blue), as encoded by the colormap.

\section{Methodology}
\label{sec:method}

The proposed architecture builds upon the framework presented in \cite{raganato2023attention}, referred to hereafter as APEAYN. Specifically, we introduce modifications to the Transformer attention mechanism by incorporating a pre-computed Gaussian function applied to the processed points.
Given two input point clouds, $X$ and $Y$, containing $n_X$ and $n_Y$ points, respectively, we concatenate the coordinates features of these point clouds separated by a separator token, SEP, of shape $1 \times 3$. Consequently, the input to the network is a matrix of shape $(n_X + 1 + n_Y) \times 3$.
In the following, without loss of generality, we consider $n=n_X=n_Y$, and thus the input matrix with shape $(2n + 1) \times 3$.

To align with the dimensionality requirements of the Transformer encoder, the two-point clouds are projected into a $d$-dimensional space using a point-wise fully connected neural network, with $d=512$ as set in the the APEAYN model.
This transformation operates independently for each point, resulting in a higher-dimensional encoding of the original 3D coordinates.
Instead, the output of the Transformer network is similarly projected back into a 3-dimensional space using a fully connected layer with learned weights.

The Transformer encoder architecture comprises multiple multi-head attention and feed-forward layers. Each multi-head attention layer consists of $h=8$ attention functions named heads. In our proposed architecture, some heads employ the conventional dot-product attention mechanism, while others utilize the Euclidean Gaussian function for attention weights. In both cases, residual attention is incorporated by adding the attention energy from the previous layer to the current layer.

The output of the network is a matrix of shape $(2n + 1) \times 3$, where each row corresponds to a point of the input mapped on the other shape.
This matrix can be split into two point clouds, $\hat{X}$ and $\hat{Y}$, once removed the SEP element. $\hat{X}$ contains the points of $X$ remapped to fit the shape of $Y$, and conversely, $\hat{Y}$ contains the points of $Y$ remapped over $X$.

\paragraph*{Dot-product attention heads.}
Given an input sequence of length $n$ and dimensionality $d$, for the $i$-th head in the $j$-th attention layer, if the head is a dot-product one, three linear projections are computed:
query $\mathbf{Q}_{j,i} \in \mathbb{R}^{n \times \tilde{d}}$,
keys $\mathbf{K}_{j,i} \in \mathbb{R}^{n \times \tilde{d}}$
and values $\mathbf{V}_{j,i} \in \mathbb{R}^{n \times \tilde{d}}$,
where $\tilde{d} = d/h$.
To incorporate positional information into the Transformer, rotary positional encoding (\textit{RoPE} \cite{su2024roformer}) is used as in \cite{raganato2023attention}, where it is shown to produce the best results with this network architecture. This method applies rotation matrices derived from the cosine and sine functions to the keys matrix.
The attention energy is then computed as:
\begin{align}
    \xi_{j,i} = softmax\left(\frac{\mathbf{Q}_{j,i}\mathbf{R}_\Theta^{\tilde{d}}\mathbf{K}_{j,i}^T}{\sqrt{\tilde{d}}} + \xi_{j-1,i}\right)
\end{align}
where $\mathbf{R}_\Theta^{\tilde{d}}$ is the block diagonal matrix with rotation matrices for each input on its diagonal. In particular, we use 
$\Theta = \{\theta_i = 10000^{-2(i-1)/\tilde{d}}, i \in [1,2,...,\tilde{d}/2]\}$.
For the input at any given $m$ position the matrix $\mathbf{R}_{\Theta, m}^{\tilde{d}}$ has the following matrices on its diagonals:
\begin{align*}
    \begin{pmatrix}
        \cos m \theta_i & -\sin m \theta_i\\
        \sin m \theta_i & \cos m \theta_i
    \end{pmatrix}
\end{align*}

\paragraph*{Gaussian attention heads.}
In Gaussian attention heads, the input is projected only in the values matrix $\mathbf{V}_{j,i} \in \mathbb{R}^{n \times \tilde{d}}$. The attention energy for each point is computed as a Euclidean Gaussian around the point:
\begin{align}
\label{eq:gaussian}
    \xi_{j,i} = softmax\left(\exp\left(-\frac{\mathbf{E}^2}{2\sigma_i^2}\right) + \xi_{j-1,i}\right)
\end{align}
where $\sigma_i$ is a fixed or learnable parameter of the network, and $\mathbf{E}$ is a $(2n+1) \times (2n+1)$ matrix and $\mathbf{E}_{p,q}$ is the Euclidean distance between the points $p$ and $q$ if they belong to the same shape, or $0$ if the two points belong to different shapes.

\begin{figure}[h]
    \centering
    \includegraphics[width=\linewidth]{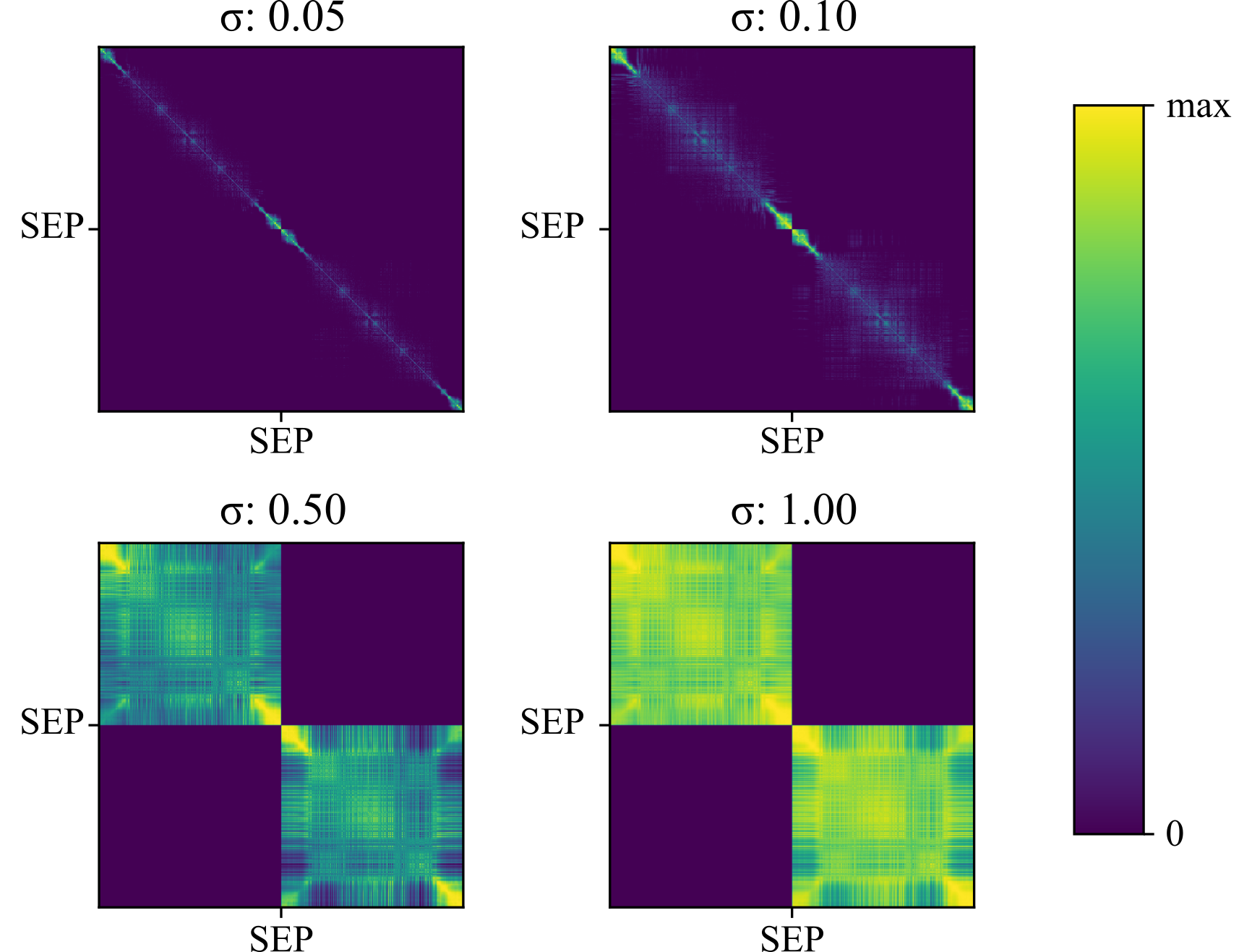}
    \caption{Attention weights of four Gaussian attention heads with different sigmas.}
    \label{fig:fixed_attention_patterns}
\end{figure}

In general, due to how the concatenation is performed, $\mathbf{E}$ will have null upper-right and lower-left quadrants as shown in Figure \ref{fig:fixed_attention_patterns}. This represents a self-attention head as all the matrix portions representing cross-shape relations are null.

Finally, the output of each attention head is computed as the weighted average of the values $\mathbf{V}_{j,i}$:
\begin{align}
    Att(\xi_{j,i}, \mathbf{V}_{j,i}) = \xi_{j,i}\mathbf{V}_{j,i}
\end{align}
The outputs of the $h$ attention heads are then concatenated and fed to a feed-forward block composed of two linear layers with ReLU activation functions.
To improve training stability a layer normalization module is placed before and after the feed-forward block.

\paragraph*{Loss computation.}
Given the two output point clouds $\hat{X}$ and $\hat{Y}$, that map the points of $X$ over the shape of $Y$ and vice versa, using the ground truth correspondence maps $\Pi_{X,Y}$ and $\Pi_{Y,X}$ we order the point clouds to respect the matching, meaning that $(x_i, y_i)\in\Pi_{X,Y}$ and $(y_i, x_i)\in\Pi_{Y,X}$ for any index $i$.
This correspondence allows to compute a simple loss as the sum of two losses $l_{X,Y}$ and $l_{Y,X}$ defined as:
\begin{align}
    l=l_{X,Y}+l_{Y,X} = ||\hat{Y}-X||_2^2 + ||\hat{X}-Y||_2^2
\end{align}
To improve the capability of the network to discriminate between which shape the points belong to, an additional mean squared error loss over the separator SEP is defined, as described in \cite{raganato2023attention}.

\section{Experimental settings and evaluation}
In this section, we outline the experimental settings, including the training and testing datasets, and define the evaluation metrics. We then present the results achieved by our models and compare them with the baseline model.

\subsection{Datasets and settings}
We begin by detailing the training data and the data augmentation methods applied. We then define the parameters of the trained models and the baseline model, followed by a description of the test data and evaluation metrics.

\paragraph*{Training data.}
Following the setup of \cite{raganato2023attention}, we utilize the same training set, which consists of 10,000 point clouds, each containing 1,000 points, extracted from the SURREAL dataset \cite{varol17_surreal}. The correspondence of the points is known by construction, as the 1,000 selected points are consistent across each shape, and this is used as ground truth during the training process.

\paragraph*{Applied augmentations.}
To enhance the model's robustness to rotations, for each input shape, we randomly apply one selected rotation from the following five possibilities: i) random rotation along all three axes within the interval $[0, 2\pi]$; ii) random rotation along the x-axis within the interval $[0, 2\pi]$; iii) random rotation along the y-axis within the interval $[0, 2\pi]$; iv) random rotation along the z-axis within the interval $[0, 2\pi]$; v) no rotation.
Furthermore, to ensure the network is independent of the order of points in the point cloud and to prevent it from learning the order, a random permutation is applied to the input points in each shape before concatenating the two shapes. This permutation is then applied to the points before computing the losses to ensure the correct use of the ground truth correspondence.

\begin{figure*}[ht]
    \centering
    \includegraphics[width=\linewidth]{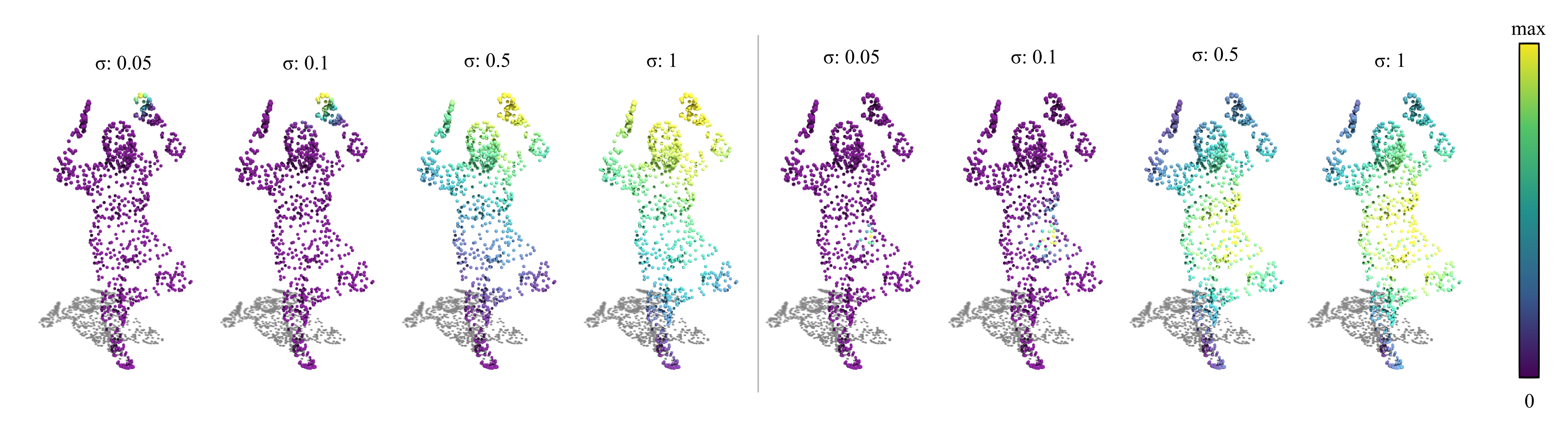}
    \caption{Fixed attention patterns provided to the model in the \textbf{4gh600} configuration plotted for two points. The first point is localized on the left hand of the shape (left), the other on the lower abdomen (right). We encode the value of the attention weights by the colormap. For each Gaussian, the value of the $\sigma$ selected is reported on the top of the visualization.}
    \label{fig:fixed_attention_patterns_shapes}
\end{figure*}

\paragraph*{Gaussian attention heads.}
We conduct our experiments using a multi-head attention Transformer encoder model with four Gaussian heads and four standard dot-product attention heads. We train two configurations:
\begin{itemize}
    \item \textbf{4gh600}: In this configuration, the sigma values are fixed based on the intuition that having different heads focus on neighborhoods of varying sizes allows the network to learn both fine and coarse details of the point cloud. The selected values for the $\sigma$ parameter in Equation \ref{eq:gaussian} are [0.05, 0.1, 0.5, 1] fixed as absolute values over the shapes after the pre-processing phase, which, as illustrated in Figure \ref{fig:fixed_attention_patterns_shapes}, approximately describe a bell curve over different regions of human shapes: the finger, the hand, the hand and forearm, and the whole arm, respectively.
    \item \textbf{4gh600.lis}: In this configuration, the sigma values are initialized to [0.05, 0.1, 0.5, 1] but are then optimized as parameters of the network.
\end{itemize}
All configurations are trained for a total of 600 epochs with batch sizes of 24 shapes (paired up into 12 couples).

\paragraph*{Baseline.}
A configuration with no Gaussian attention heads, identified by the code \textbf{0gh600}, is trained for 600 epochs with a batch size of 24. This configuration reproduces the APEAYN model with a reduced number of epochs, from 5000 to 600, to have a fairer comparison to the other trained models.

\paragraph*{Test data.}
The models are tested on the shape matching task using the Faust1k dataset \cite{bogo2014faust}. This dataset comprises 10 subjects in 10 different poses, represented by the same mesh, totaling 100 shapes with 1,000 points each. Additionally the models are tested on a noisy version of the same dataset, as done in \cite{raganato2023attention}, to assess the robustness to noise of the methods. The noise on the test set is extracted from a normal distribution $N(0,0.01)$ and added on every point in the dataset.

\paragraph*{Metrics.}
The evaluation of the results follows the setup in \cite{trappolini2021shape,raganato2023attention}.
From the Faust1k dataset, 100 random pairings are selected as a test set. 
Each inference on a pair produces two output point clouds, $\hat{X}$ containing the points of $X$ remapped over the shape of $Y$, and $\hat{Y}$ remapping $Y$ over $X$.
These constitute the data to compute two matchings, $\pi_{X,Y}$ and $\pi_{Y,X}$, which map the points of $X$ to $Y$ and vice versa, respectively.
To select which matching is to be computed, for each inferred point cloud we compute the chamfer loss against the ground truth.
The couple that produces the lower chamfer error is used to compute the matching.
In particular, in the case $\hat{X}$ and $Y$ produce the lower loss value, for each point $x$ in the source point cloud $X$ we select the matching point $y_x$ in $Y$ as the point in $Y$ closest to $\hat{x}$, mapping of $x$ over the target shape $Y$.
The error value is then defined as the geodesic distance between $y_x$ and $y$, point in $Y$ such that $(x,y)\in\Pi_{X,Y}$, ground truth matching.
It is important to note that the network processes the mesh as a point cloud, retaining no information on connectivity. The geodesic distances are used solely as an evaluation metric.

The implementation of the training process, the evaluation and the models used for the experiments can be found at this repository: 
\href{https://github.com/ariva00/GaussianAttention4Matching}{github.com/ariva00/GaussianAttention4Matching}.

\subsection{Results}
The results on the Faust1k dataset are presented in Table \ref{tab:results}. For each trained model, two checkpoints are listed: one at 600 epochs and another at the epoch that produced the smallest loss.

\begin{table}[ht]
\centering
\caption{Comparison of the trained models results. For each model, we report the average geodesic error on the clean dataset (F1k error) and on the noisy dataset (F1k$_{noise}$ error), the number of epochs of its learning, and the number of parameters. For the models we propose and for the baseline, we consider both the version obtained after 600 epochs and the one that reaches the best minimization of the loss (denoted as \emph{best.}). }
\label{tab:results}
\resizebox{\linewidth}{!}{
\begin{tabular}{@{}lllrr@{}}
\toprule
Model       & F1k error & F1k$_{noise}$ error & Epochs & Parameters \\ \midrule
0gh600                  & 0.0231    & 0.0379    & 600 & 19.25M \\
4gh600                  & 0.0227    & 0.1234    & 600 & 17.68M \\
4gh600.lis              & 0.0198    & 0.1264    & 600 & 17.68M \\ 
best.0gh600             & 0.0206    & 0.0365    & 578 & 19.25M \\
best.4gh600             & 0.0171    & 0.1358    & 576 & 17.68M \\
best.4gh600.lis         & 0.0223    & 0.1578    & 574 & 17.68M \\
\textit{APEAYN} \cite{raganato2023attention} & \textit{0.0124} & \textit{0.0282} & \textit{5000} & 19.25M \\ \bottomrule
\end{tabular}
}
\end{table}

It is evident that with the same number of epochs, the model utilizing Gaussian attention heads produces better results and reduces the number of parameters. The error is approximately 15\% less than that of the baseline model.
The improved architecture achieves an average matching error of only 160\% of the error of the APEAYN model in just 12\% of the training time, the latter being trained for 5K epochs. Moreover, the 0gh600 model produces an error that is 200\% of that of the APEAYN model.
Interestingly, the errors recorded by the models at the epoch with the best training loss do not rank in the same order, indicating a potential instability in the loss descent.

The resulting four $\sigma$ values of the 4gh600.lis model are $0.03$, $0.09$, $0.22$, and $0.87$. The distribution of these parameters after training suggests that the initial values were too large, but adopting multiple scales is a beneficial approach.

\paragraph*{Noisy data.} It is possible to notice how the performances of the models with gaussian heads drastically decrease when faced with noisy data. The error on the noisy version of the Faust1k dataset is more than five times the error on the clean version for the 4gh600 and 4gh600.lis models, while it is only 65\% more in the case of the 0gh600 model.
This is due to the nature of the selected neighborhood in the gaussian heads, which is based solely on euclidean distances and is inherently very susceptible to noise.
To mitigate this problem the \textbf{4gh600.lis.noise} model has been introduced, it is trained identically to the 4gh600.lis model, with the addition of noise in the training set. The noise is injected on 50\% of the points and extracted from a normal distribution $N(0, 0.02)$, this additional data augmentation should also help in avoiding overfitting issues.
The results, reported in Table \ref{tab:results_plus_noise} shows that the model is indeed more robust to noise with respect to the 4gh600.lis model but the performances on the clean dataset also decrease to the point it is slightly worse than the baseline, revealing that this is not the best solution to the problem and that further studies need to be conducted.

\begin{table}[t!]
\centering
\caption{Comparison of the trained models results with the additional model trained with noisy data 4gh600.lis.noise. For each model, we report the average geodesic error on the clean and noisy datasets}
\label{tab:results_plus_noise}
\begin{tabular}{@{}lllrr@{}}
\toprule
Model       & F1k error & F1k$_{noise}$ error\\
\midrule
0gh600                  & 0.0231    & 0.0379\\
4gh600                  & 0.0227    & 0.1234\\
4gh600.lis              & 0.0198    & 0.1264\\
\textbf{4gh600.lis.noise}        & 0.0249    & 0.0395\\
\bottomrule
\end{tabular}
\end{table}

\begin{figure*}[ht]
    \centering
    \includegraphics[width=\linewidth]{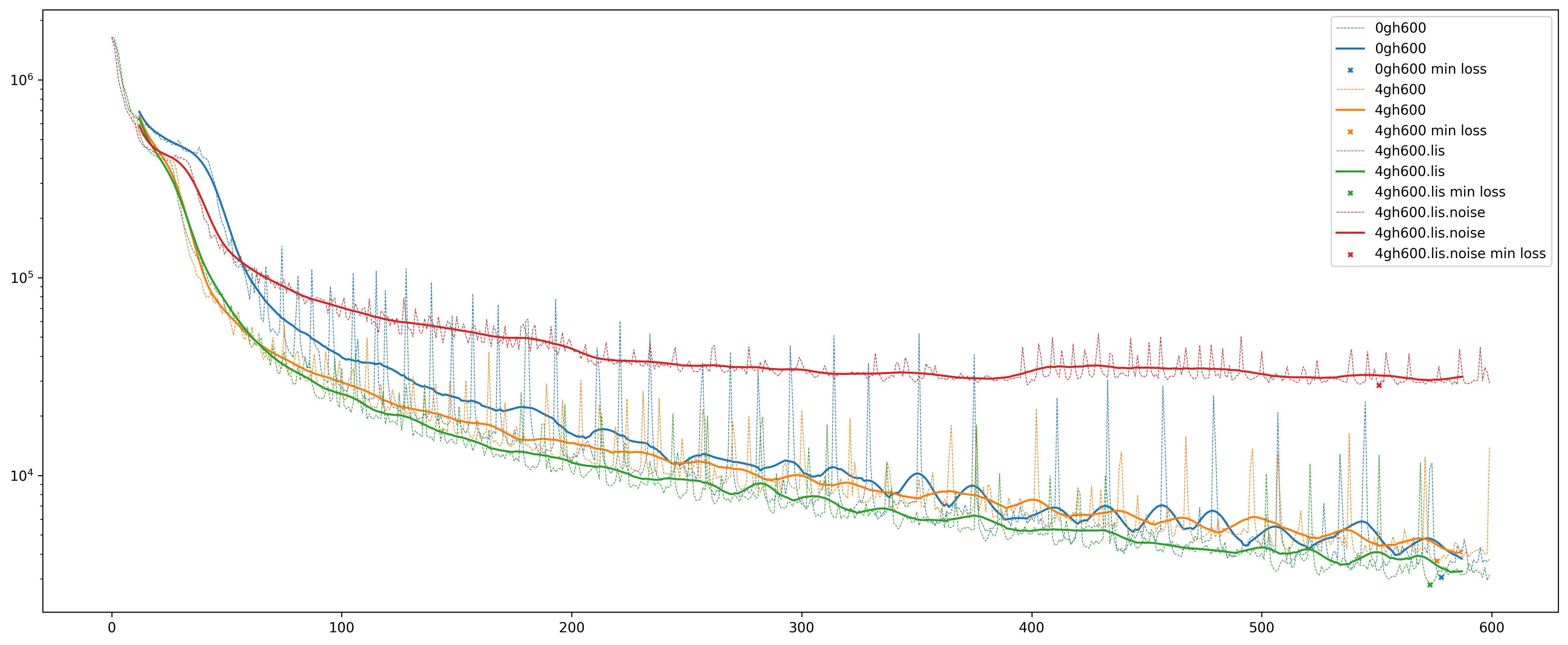}
    \caption{Training loss through the epochs of the models. The curves are reported unfiltered (dashed line) and smoothed (solid line) for a clearer visualization.}
    \label{fig:training_loss}
\end{figure*}

\section{Analysis and Ablation}
To better understand the results obtained by the models, we plot the loss descent through the epochs in Figure \ref{fig:training_loss}. It is evident that all the loss curves exhibit significant noise, yet the spikes in the Gaussian heads models appear less intense, particularly in the 4gh600.lis model that optimizes the sigma values. This suggests that the training loss is not an ideal metric for identifying the best model, indicating the need for a better validation method.

The smoothed version of the curves shows that the 4gh600.lis model consistently produces lower loss values, suggesting that the injection of fixed Gaussian attention improves the training process as expected. Notably, in the initial epochs, both the 4gh600.lis and 4gh600 models demonstrate a smoother and faster descent.
All models still exhibit improvement close to epoch 600, indicating that additional training could achieve better results.

The model trained with the noisy dataset results in loss values higher than all the other models along all the epochs but the first ones, where it registers loss values lower than the baseline. This is not unexpected as the loss values are computed on noisy data that is intrinsically more complex and challenging. It is however interesting to notice how the minimum loss epoch occurs earlier with respect to the other models and the plateau seems to be more pronunced. This suggests that the addition of noise in training helps the method adapt to noisy situations but also slows down the loss decrease, at least with the selected noise distribution.

\begin{figure}[ht]
    \centering
    \includegraphics[width=\linewidth]{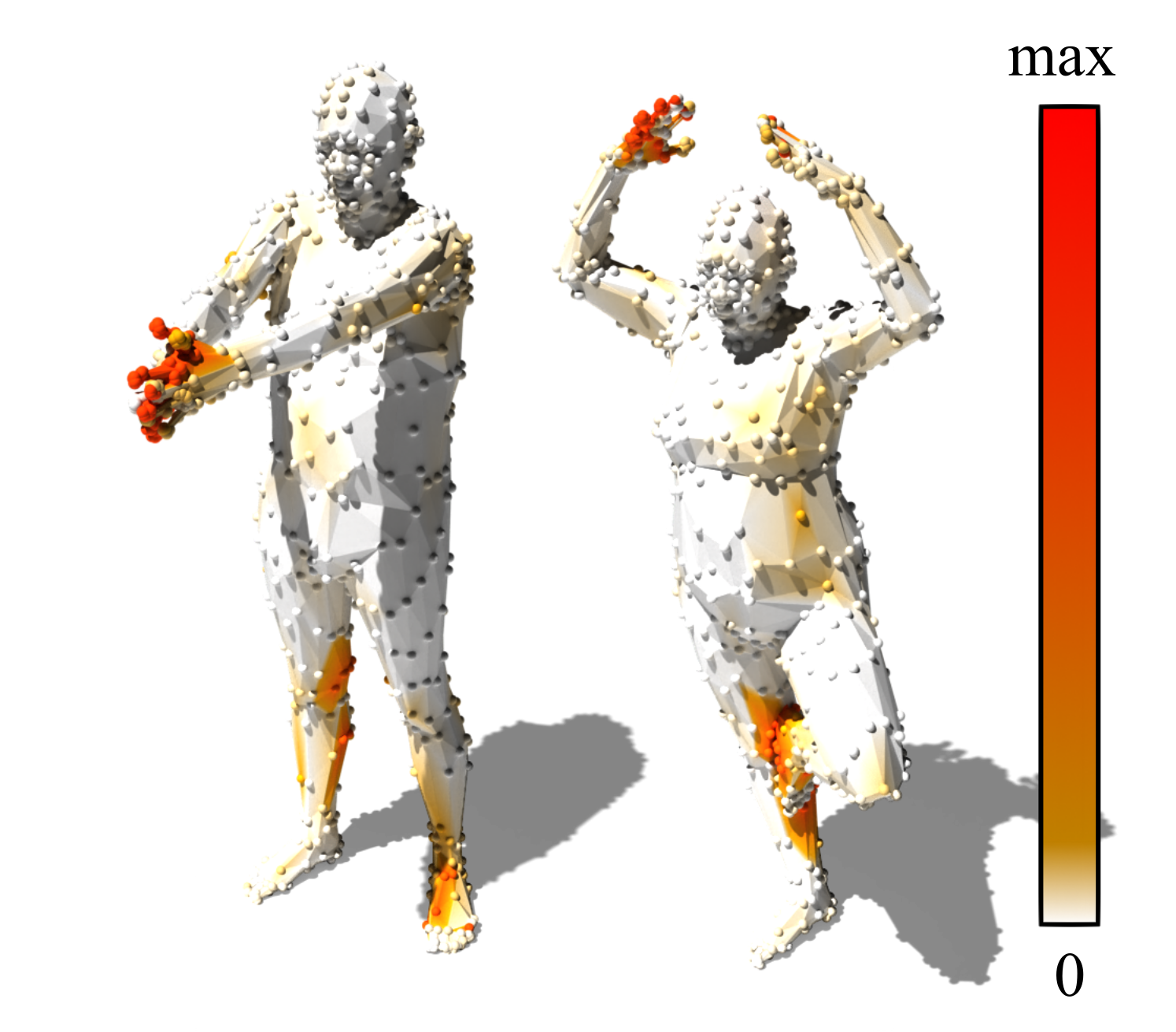}
    \caption{Error values for a pair of point clouds obtained by the 4gh600 model. The colormap encodes the error: zero errors are white, while larger errors are darker.}
    \label{fig:error_patterns}
\end{figure}

Figure \ref{fig:error_patterns} illustrates a pair of shapes on which the 4gh600 model records one of the highest matching errors on the test set. The errors are localized around the hands and between the left foot and the right knee. This is due to the close proximity of points from different regions; the model lacks information about point connectivity and thus cannot distinguish between points that are close in 3D space but far apart on the underlying manifold.
For visualization purposes, we render the mesh under the point clouds with the same colors to better visualize the error areas.

\begin{figure*}[ht]
    \centering
    \includegraphics[width=\linewidth]{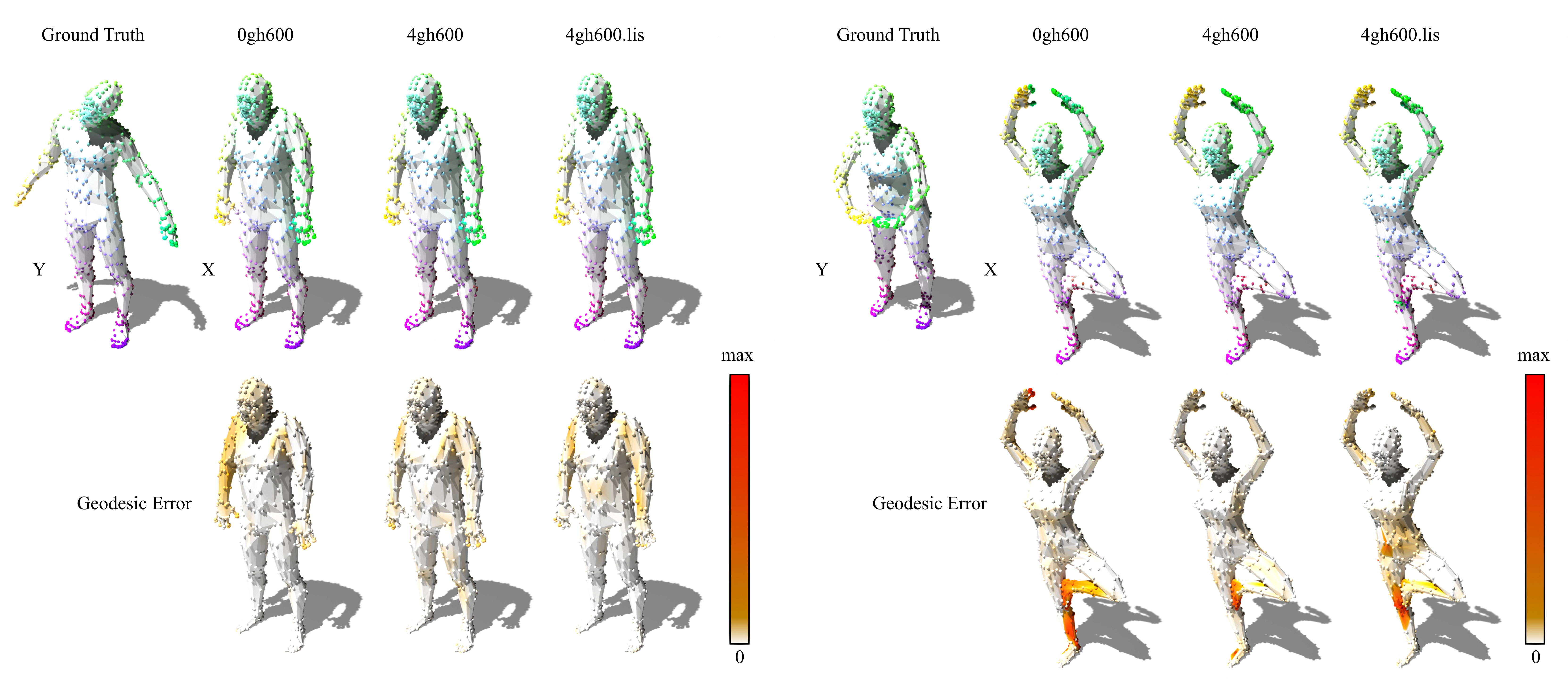}
    \caption{Qualitative comparison of the matching results on two pairs of shapes with the trained models. On the left, an optimal case, on the right, a more challenging one. In the first row, we report the estimated correspondence using color coding. In the second row, we depict the geodesic error encoded by the colormap.}
    \label{fig:matching}
\end{figure*}

Figure \ref{fig:matching} compares the matchings produced by the trained models. The example on the left shows an optimal case where the lack of connectivity information does not cause significant errors. Conversely, the right side shows another example of a critical situation, similar to Figure \ref{fig:error_patterns}.
Overall, our architectures demonstrate lower and more localized errors than the 0gh600 model, although the critical regions are consistent across all models.

\begin{table}[ht]
\centering
\caption{Comparison of the trained models under random rotations and random permutations applied to the input point clouds.}
\label{tab:results_rot_perm}
\resizebox{\linewidth}{!}{
\begin{tabular}{@{}llll@{}}
\toprule
\multirow{2}{*}{Model}  & \multirow{2}{*}{F1k error}    & Random  & Random        \\
                        &                               & Rotation & Permutation \\ \midrule
0gh600      & 0.0231    & 0.0231  &  0.0185      \\
4gh600      & 0.0227    & 0.0227  &  0.0216      \\
4gh600.lis  & 0.0198    & 0.0207  &  0.0140      \\ 
best.0gh600      & 0.0206    & 0.0213  &  0.0183      \\
best.4gh600      & 0.0171    & 0.0176  &  0.0145      \\
best.4gh600.lis  & 0.0223    & 0.0232  &  0.0127      \\ 
\textit{APEAYN} \cite{raganato2023attention} & \textit{0.0124} & \textit{0.0127} & \textit{0.0112}\\ \bottomrule
\end{tabular}
}
\end{table}

Table \ref{tab:results_rot_perm} showcases the performance of the models when random rotations and permutations are applied to the input point clouds. It is evident that the trained models do not suffer significant degradation due to these transformations, reflecting that the augmentations during training are sufficient to ensure robustness.

\begin{table*}[ht]
\centering
\caption{Results of the ablation study on the attention heads. Each column AblationN reports the results with the N-th attention head masked out. In brackets, we report the sigma values for the Gaussian heads.}
\label{tab:ablation_heads}
\resizebox{\textwidth}{!}{%
\begin{tabular}{@{}llllllllll@{}}
\toprule
Model                       & Full model & Ablation1 & Ablation2 & Ablation3 & Ablation4 & Ablation5 & Ablation6 & Ablation7 & Ablation8 \\ \midrule
\multirow{2}{*}{0gh600}     & -         & cross     & self      & self      & cross     & self      & self      & self      & self \\
                            & 0.0231    & 0.1001    & 0.2585    & 0.1570    & 0.0458    & 0.0485    & 0.2851    & 0.2026    & 0.1896 \\ \midrule
\multirow{2}{*}{4gh600}     & -         & self      & cross     & cross     & self      & self [0.05]     & self [0.10]      & self [0.50]     & self [1.00] \\
                            & 0.0227    & 0.0513    & 0.1791    & 0.1576    & 0.0965    & 0.2950    & 0.3333    & 0.3807    & 0.2891 \\ \midrule
\multirow{2}{*}{4gh600.lis} & -         & self      & cross     & cross     & self      & self [0.03]     & self [0.09]      & self [0.22]     & self [0.87] \\
                            & 0.0198    & 0.0423    & 0.1179    & 0.3273    & 0.1029    & 0.2858    & 0.3930    & 0.3597    & 0.3370 \\ \midrule
\multirow{2}{*}{APEAYN 
\cite{raganato2023attention}}      
                            & -         & self      & self      & cross     & self      & self      & self      & cross     & self \\
                            & 0.0124    & 0.0282    & 0.2462    & 0.2254    & 0.3166    & 0.0210    & 0.0156    & 0.1300    & 0.0476 \\ \bottomrule

\end{tabular}%
}
\end{table*}

\subsection{Ablation study}
In this section we report and analyze the results of two ablation studies conducted on the models:
\begin{itemize}
    \item \textbf{Heads ablation.}
    To better understand the contribution of each attention head and the importance of the information it carries for the matching problem, for every model we completely mask out the attention of each head, one at a time, and compare the results on the test set with those of the complete model.
    Table \ref{tab:ablation_heads} reports the results of the different models with each head masked out.
    \item \textbf{Layers ablation.}
    To identify the specific layers where the Gaussian information is most impactful, we train six additional models for 100 epochs. Each of these models uses only one multi-head attention layer with fixed attention weights, while the other layers are kept with the standard dot-product attention.
    The results of the tests are shown in Table \ref{tab:ablation_layers}.
\end{itemize}

\begin{figure}[ht]
    \centering
    \includegraphics[width=\linewidth]{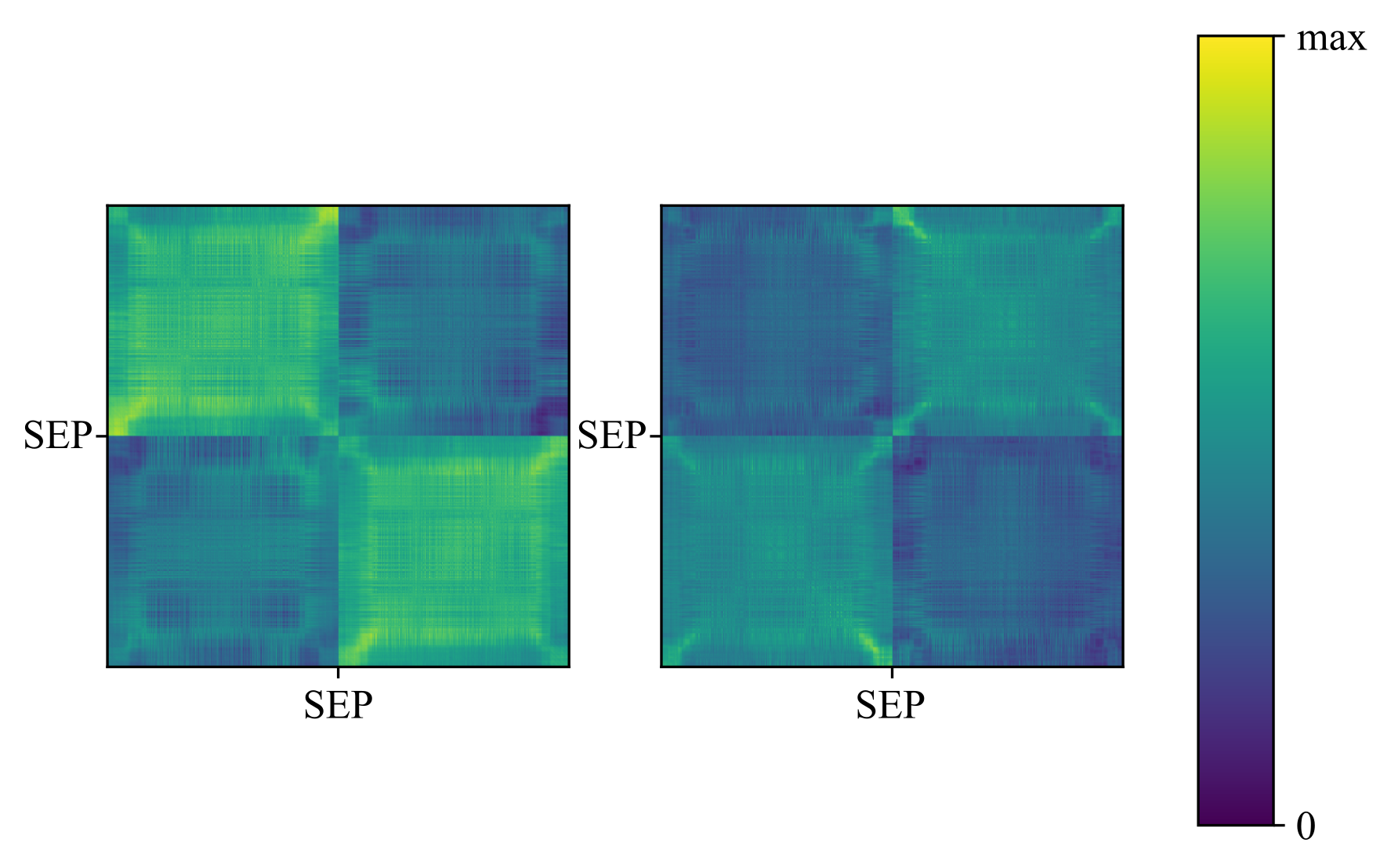}
    \caption{Attention weights of two dot-product attention heads of the 4gh600 model in the final attention layer. The left head shows a self attention pattern, the right one shows a cross-attention pattern.}
    \label{fig:learned_attention_patterns}
\end{figure}

Before delving into the analysis, it is essential to define what self-attention and cross-attention heads mean in the context of this work.
Conventionally, a self-attention head computes attention between the input and itself, while a cross-attention head computes attention between two different input sequences.
For the scope of this study, we define:
\begin{itemize}
\item Self-Attention Heads: Heads that produce attention weight matrices with higher values in the upper-left and lower-right quadrants than in the upper-right and lower-left quadrants. These heads provide more information regarding the relationships between points within the same shape. In Figure \ref{fig:learned_attention_patterns} the left head is a self-attention head.
\item Cross-Attention Heads: Heads that produce attention weight matrices with higher values in the upper-right and lower-left quadrants. These heads focus more on the relationships between points from different shapes. The right head in \ref{fig:learned_attention_patterns} classifies as cross-attention head.
\end{itemize}

By masking out one head at a time and evaluating the impact on performance, we can gain insights into the specific roles and importance of self-attention and cross-attention heads in the matching process.

\paragraph*{Heads ablation.}
In Table \ref{tab:ablation_heads}, the results of the ablation study on the heads are presented. Each column shows the matching error with the corresponding head masked out, as well as the type of head, whether it was self-attention or cross-attention.

The \textbf{0gh600} model has 2 out of 8 heads designated as cross-attention heads. The ablation results indicate that the model relies primarily on the self-attention heads, though not uniformly. The error among the self-attention heads ranges from 200\% to 1200\% of the baseline error. No head is particularly redundant or superfluous, as even the smallest error recorded in the ablation tests is almost double the error registered by the full model.

The \textbf{4gh600} model heavily relies on the fixed Gaussian heads (heads from 5 to 8), as their removal results in the highest errors. Specifically, the ablation of each Gaussian head shows an error ranging from 1300\% to 1700\% of the complete model error. The other errors range from double to ten times the full model error, indicating no significant information redundancy across the heads. Interestingly, among the dot-product heads, the model relies most on the cross-attention ones, contrasting with the reliance pattern observed in the 0gh600 model. The reliance on fixed attention heads suggests that valuable information has been present since the beginning of epoch one, unlike the randomly initialized heads.

The \textbf{4gh600.lis} model behaves similarly to the 4gh600 model, relying heavily on the Gaussian heads and showing no particular signs of redundancy. The main difference is found in the error for the masked head 3. In both models, this head is a cross-attention head, but in the model with learned sigma values, masking this head produces significantly worse results, even worse than masking head 5, one of the Gaussian heads.

In addition to the trained models, the \textbf{APEAYN} model is also used for the ablation study. The results reveal some redundancy, particularly in the ablation of head 6, which shows very little error difference. Overall, the errors introduced by removing one head at a time are smaller than those of the other models but still significant. Unlike the other models, the cross-attention specialized heads appear more crucial for the model's performance than most self-attention heads.

\paragraph*{Layers ablation.}

Table \ref{tab:ablation_layers} presents the results of the ablation study on the attention layers. Six models were trained for 100 epochs each, with only one of the six attention layers utilizing the Gaussian heads. 
The best results were achieved by the Layer4 model, which employed Gaussian attention in its fourth layer. Overall, the network benefits more from the additional information when it is provided in the deeper layers. This suggests that deeper layers are more effective at integrating and leveraging the fine-grained details captured by the Gaussian heads, thereby enhancing the model's performance.

\begin{table}[!ht]
\centering
\caption{Results of the ablation study on the attention layers. Each row LayerN reports the results of the model with the N-th attention layer that uses four fixed Gaussian attention heads, while all the other layers present full standard dot-product attention heads. Each model is trained for 100 epochs.}
\label{tab:ablation_layers}
\begin{tabular}{@{}ll@{}}
\toprule
Model       & F1k error             \\ \midrule
Layer0      & 0.0945    \\
Layer1      & 0.1207    \\
Layer2      & 0.0740    \\ 
Layer3      & 0.0694    \\
Layer4      & 0.0497    \\
Layer6      & 0.0901    \\ 
\textit{4gh100} & \textit{0.0360} \\\bottomrule
\end{tabular}
\end{table}

\subsection{Limitations}
In this work, we focused on the shape matching task between human shapes only. Moreover, the tests were performed on a dataset that exclusively comprises point clouds of the same cardinality. 
With our work we aim to analyze in depth which type of geometric information enables a transformer architecture to achieve state-of-the-art results. For this reason, in our experiments, we limit our comparison to the most recent data-driven solution for shape matching based on transformers.
Extensive testing on diverse and more challenging datasets, including both human and non-human shapes, is required to fully assess the applicability, generalizability, and potential drawbacks of the proposed methodology, as well as a broader comparison to state-of-the-art methods.
The performaces on noisy data of the 4gh600.lis.noise show how the methodology could work but does require further studies to limit the matching deterioration on the clean dataset.
\section{Conclusions}
\label{sec:conclusions}
In this work, we showed that in a Transformers-based approach to the shape matching task, substituting dot-product attention heads with Gaussian attention heads in the Transformer architecture significantly accelerates the training process, particularly during the initial stages. The enhanced architecture not only exhibited improved performance metrics but also showed a smoother loss descent.
The robustness to noise can be improved via injection of noise in the training set, however it causes a performance decrease when performed in the current configuration, thus requiring further studies.
Furthermore, our ablation study highlighted the importance of the fixed attention heads within the architecture and the optimal depth at which Gaussian information can be effectively injected.

While our testing has not been exhaustive, we believe that with further refinement and an enhanced training process, a similar architecture that leverages localized Gaussians could further reduce training time and yield more stable results. In future work, we plan on validating these findings across a broader range of datasets \cite{melzi:2019:shrec19,DykeShrec20} to confirm the robustness and versatility of our approach.

\section*{Acknowledgments}
This work was partially supported by the MUR for REGAINS, the Department of Excellence DISCo at the University of Milano-Bicocca, the PRIN project GEOPRIDE Prot.\ 2022-NAZ-0115, CUP
H53D23003400001, and by the NVIDIA Corporation with the RTX A5000 GPUs granted through the Academic Hardware Grant Program to the 
University of Milano-Bicocca for the project ``Learned representations for implicit binary operations on real-world 2D-3D data''.
\bibliographystyle{eg-alpha-doi} 
\bibliography{egbibsample}       



\end{document}